\documentclass{article}
\usepackage{amsmath}
\usepackage{spconf}
\usepackage{graphicx}
\usepackage{multirow}
\usepackage[flushleft]{threeparttable}
\usepackage{xcolor}
\usepackage{enumitem}
\usepackage{adjustbox}
\usepackage{cite}

\newcommand{\cblue}[1]{{\color{black} #1}}
\newcommand{\crblue}[1]{{\color{black} #1}}

\title{\vspace{-5mm} Language Agnostic Data-Driven Inverse Text Normalization \vspace{-3mm}}

\name{Szu-Jui Chen$^{1}${\sthanks{Work performed during an internship at Meta Platforms, Inc.}\sthanks{Equal contribution.}},
Debjyoti Paul$^{2}$\footnotemark[2],\textit{Yutong Pang}$^{2}$,  \textit{Peng Su}$^{2}$, \textit{Xuedong Zhang}$^{2}$
\vspace{-4mm}}
\address{Center for Robust Speech Systems (CRSS), University of Texas at Dallas, USA$^{1}$,\\
Meta AI, USA$^{2}$ \vspace{-6mm}}

\begin{document}
\maketitle
\begin{abstract}
With the emergence of automatic speech recognition (ASR) models, converting the spoken form text (from ASR) to the written form is in urgent need. This inverse text normalization (ITN) problem attracts the attention of researchers from various fields. Recently, several works show that data-driven ITN methods can output high-quality written form text. Due to the scarcity of labeled spoken-written datasets, the studies on non-English data-driven ITN are quite limited. In this work, we propose a language-agnostic data-driven ITN framework to fill this gap. Specifically, we leverage the data augmentation in conjunction with neural machine translated data for low resource languages.
Moreover, we design an evaluation method for language agnostic ITN model when only English data is available. Our empirical evaluation shows this language agnostic modeling approach is effective for low resource languages while preserving the performance for high resource languages.
\end{abstract}
\vspace{-1.5mm}
\vspace{-4mm}
\section{Introduction}
\vspace{-3mm}
Recent progress in automatic speech recognition (ASR) technologies brings world-wide adoption to use voice as a source of input to interact and communicate with digital environments. Moreover, ASR systems are now expanding language horizons to bring same or better experience to native language speaker around the world. For communication and better readability of spoken form output from an ASR system, it needs to be paired with an inverse text normalization (ITN) system to produce corresponding written form texts.

\cblue{Converting spoken form texts to written form is not an easy task.}
For example, the spoken form \textit{twenty twenty} can be written as (a) \textit{2020} to express the year/number, (b) \textit{20:20} for time, (c) \textit{20/20} for eye-vision, (d) {20-20} for a score in game or a cricket match. Taking another example of \textit{ten to twelve}, which can be written as (a) \textit{10-12} as a cardinal number range, (b) 10:00-12:00 as time range, (c) \textit{11:50 am/pm} as a time instance, etc. It is evident from these examples that the speaker contexts could help in determining the correct written form which is hard to abide by a rule-based system. Therefore data-driven based systems are gaining attention of researchers \cite{pusateri2017mostly, lai2021unified, sunkara2021neural, antonova2022thutmose} to improve ITN systems, dub as DD-ITN models.

\cblue{Dataset of spoken-written text} pairs are used to train these neural network based DD-ITN models. Obtaining spoken-written text pairs that covers diverse ITN entities such as cardinals, ordinals, date-time, money, fractions, decimals, address, metrics, email, URL, abbreviation etc. is hard but doable for high-resource languages such as English.
\cblue{In addition, the written representation of same entity can vary across languages, e.g., 3:30 pm is commonly represented as 15h30 in French.}
\cblue{The challenge}
grows multi-fold while expanding to more languages, especially low-resource ones where it is hard to obtain spoken-written text pairs dataset.

\crblue{
We discovered that separating the ITN component from the end-to-end ASR system allows for independent performance improvements using a relatively large text only dataset. Moreover, our in-house on-device spoken-form ASR systems have achieved a significant reduction in word-error-rate (WER) of $>$10\%, a reduction in real-time-factor (RTF) of $>$8\%, and lower memory consumption. This has led us to propose using the ITN as a low-latency post-processing service. In this study, we aim to internationalize (i18n) and expand the language capabilities of ITN models by addressing the problem of data scarcity and presenting a unified, language-agnostic ITN model that supports multiple languages.} The main contributions of this work are as follows:

\vspace{-3mm}
\begin{itemize}[leftmargin=*]
    \item We propose a text normalization method for English that transforms written form texts to spoken form texts. Unlike conventional text normalization system, our data augmentation system generates more possible variants of spoken forms; which can help \cblue{build} robust ITN system.\\ \vspace{-7mm}
    \item \cblue{We propose to apply neural machine translation for internationalization of the ITN models, which can be considered as a knowledge distillation approach.} We use neural machine translation on English spoken-written text pairs to generate spoken-written pairs on target languages; and it helps ITN expanding to more languages.\\ \vspace{-7mm}
    \item We present a language agnostic data-driven ITN model that supports inverse normalization of spoken form texts for 12 languages. \cblue{We also present a study of system design choices in our experiment section.}\vspace{-1mm}
\end{itemize}




\begin{figure*}[!t]
    \centering
    \vspace{-10mm}
    \includegraphics[width=0.85\linewidth]{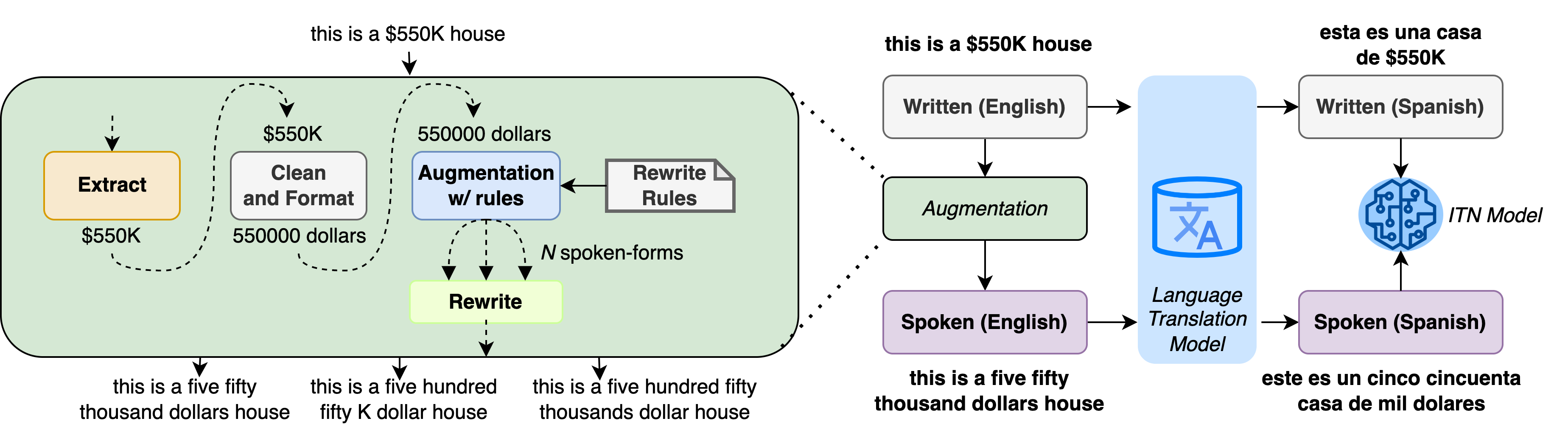}
    \vspace{-2mm}
    \caption{Multilingual data generation using rule based text normalization system and machine translation model.}
    \label{fig:pipeline}
    \vspace{-3mm}
\end{figure*}

\vspace{-7mm}
\section{Methodology}
\label{sec:method}
\vspace{-3mm}
\subsection{Data Augmentation}
\vspace{-2mm}
To train a language agnostic ITN system, we need spoken-written text pairs for every language. However, there are not much public available resources having spoken-written text pairs in English, not to mention other languages. Hence, we propose a two-step data augmentation method to generate spoken-written text pairs for multiple languages. First, we generate spoken form texts from highly available written text resources in English. Then, we use machine translation to generate text pairs for other languages.\vspace{-10mm}\\
\subsubsection{Enhanced Text Normalization}\vspace{-2mm}
Traditional text normalization (TN) used in text-to-speech (TTS) systems generates fixed variation of spoken forms (often with rule-based approach) that conforms to the verbalizer standard for any given written text with TN-ITN entities. As mentioned previously, spoken forms lack full information about the subject; hence, it is important to cover more spoken variations and alternatives. With \cblue{abundant} written text resources in hand, \crblue{we apply our specialized text normalization augmentation system that generates many possible spoken variations; statistically 22$\times$ more diverse than conventional TN system. See Table~\ref{table:aug} for a few examples and we recommend readers to check the Appendix Sec. 6.1 in our extended version of paper for details \cite{chen2023language}.}

\begin{table}[h]
\vspace{-6mm}
\centering
\caption{Examples of generated spoken form using conventional TN system and our data augmentation system}
\label{table:aug}
\vspace{1mm}
{\scriptsize
\centering
\begin{tabular}{l|l|l}
\hline
\textbf{Written Text} & \textbf{Spoken Text from}  & \textbf{Spoken Text from } \\
\textbf{Input} & \textbf{Conventional TN}  & \textbf{Enhanced TN System} \\ \hline
6:15 am           & \begin{tabular}[c]{@{}l@{}}six fifteen a m \\  ~\end{tabular}    & \begin{tabular}[c]{@{}l@{}}six fifteen a m\\ six fifteen in the morning\\ six fifteen\\ six past fifteen a m\\ quarter past six a m \\ quarter past six morning \\ six and quarter a m\end{tabular}
\\\hline
\$1.20           & \begin{tabular}[c]{@{}l@{}} one dollar and twenty cents \\  ~\end{tabular}    & \begin{tabular}[c]{@{}l@{}} one dollar and twenty cents\\ one dollar twenty cents \\one dollar two zero cents\\ one point two zero dollars \\ a dollar twenty cents\end{tabular}
\\\hline
\end{tabular}
\vspace{-1mm}
}
\vspace{-8mm}
\end{table}

\subsubsection{Multilingual spoken-written text pairs}
\vspace{-2mm}
\crblue{We propose using neural machine translation (NMT) models to generate spoken-written text pairs in target languages for which we do not have adequate pairs. However, we have found that the outputs of the NMT model do not always meet our criteria for quality. To ensure the quality of our spoken-written text pairs, we have implemented the following measures:
(a) {\em Spoken/Written Mismatch:} Discarding translated texts that have mismatches between written and spoken forms, (b) {\em Word Error Rate (WER):} Strictly adhering to the Word Error Rate (WER) metric for selecting non-ITN text segments,
(c) {\em Target Language Conformity:} Ensuring conformity between the source and target languages, and filtering out any malformed or incorrect translations with input from linguists, like {\em 801 $\rightarrow$ eight o one [\textit{en}] $\neq$ otto o uno [\textit{it}]}, etc. We provide translation accuracy in Table~\ref{tab:multilingual}, the pictorial position of translation module in the pipeline in Fig.~\ref{fig:pipeline}, and a few translation examples in Table~\ref{table:nmt_examples} for reference.}

\begin{table}[t]
\vspace{-5mm}
\centering
\caption{Examples of data augmentation with machine translation models for French [\textit{fr}], Italian [\textit{it}], Spanish [\textit{es}].}
\label{table:nmt_examples}
\vspace{1mm}
{\scriptsize
\centering
\begin{tabular}{l|l|l}
\hline
\textbf{Form} & \textbf{Text in English}  & \textbf{Translated text} \\ \hline
{\em spoken}          & \begin{tabular}[c]{@{}l@{}}Historical average for \\January is thirty one\\ degrees.\end{tabular} & \begin{tabular}[c]{@{}l@{}} La moyenne historique de janvier \\est de trente et un degrés [\textit{fr}]\\La media storica di gennaio \\è di trentuno gradi. [\textit{it}]\\
La media histórica de enero \\es de treinta y un grados. [\textit{es}] \\  \end{tabular} \\ \hline
{\em written}          & \begin{tabular}[c]{@{}l@{}}Historical average for \\January is 31 degrees.\end{tabular} & \begin{tabular}[c]{@{}l@{}} La moyenne historique pour janvier \\est de 31 degrés. [\textit{fr}]\\La media storica di gennaio \\è di 31 gradi. [\textit{it}]\\
La media histórica de enero \\es de 31 grados. [\textit{es}] \\ \end{tabular} \\ \hline
\end{tabular}
\vspace{-6mm}
}
\end{table}

\vspace{-4mm}
\subsection{Model Architecture}
\label{sec:model_arch}
\begin{figure}[b]
    \centering
    \vspace{-4mm}
    \includegraphics[width=\linewidth]{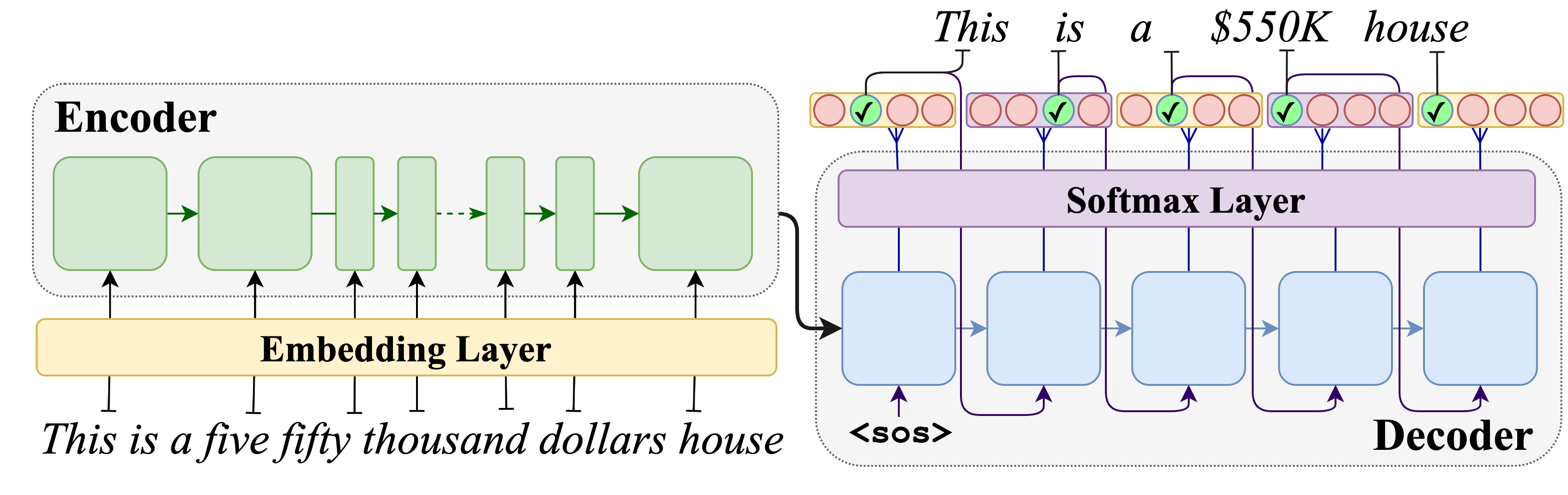}
    \vspace{-8mm}
    \caption{Encoder-Decoder model architecture for ITN.}
    \label{fig:model_arch}
\end{figure}

\vspace{-1mm}
For our ITN task, it can be seen as a sequence-to-sequence (Seq2Seq) problem, which turns a sequence in one domain (e.g., spoken domain) to a sequence in another domain (e.g., written domain). For this Seq2Seq problem, we employ the Encoder-Decoder architecture (Fig.~\ref{fig:model_arch}) to solve it. More specifically, two types of Encoder-Decoder model are investigated in this work: the LSTM-based Seq2Seq model and the Transformer model.

With the sequence data, a natural idea is to use the recurrent neural network (RNN), and Long Short-Term Memory (LSTM) \cite{hochreiter1997long} is the first choice among Recurrent Neural Networks (RNNs). In our LSTM-based Seq2Seq model, we utilize the LSTM in both the encoder and decoder. In order to encode the input sequence better, we employ bi-directional encoder to consider both the left context and right context in the sequence. Also, in the decoder side, we use attention mechanism to derive a context vector that captures the relevant source-side (encoder-side) information to help prediction. Initially, this type of Seq2Seq model is used for translation task. For more details about the model, we refer readers to paper \cite{luong2015effective}.

For the Transformer model, we employ the original model from \cite{vaswani2017attention}. Its encoder and decoder are composed of stacked modules (or layers), and each module mainly consists of multi-head attention and feed forward networks. The attention mechanism will take the whole sequence into account by learning weights for input tokens in the encoder. In the decoder, the masked attention mechanism is applied to predict the next token based on the previous tokens.

For fair comparison, we experiment with both models with similar parameter sizes.
The details of the model parameter can be found in Table~\ref{tab:model_para}.

\begin{table}[t]
\centering
\vspace{-6mm}
\caption{Model Parameters.}
\vspace{1mm}
\label{tab:model_para}
{\scriptsize
\centering
\begin{tabular}{l|l|l}
\hline
 & \textbf{LSTM}  & \textbf{Transformer} \\ \hline
{\em No. of Parameter }          & 19.98M & 19.67M  \\ \hline
{\em Encoder Layer}          & 2 & 4 \\
{\em Decoder Layer}          & 2 & 4 \\
{\em Hidden Size}          & 256 & 256 \\
{\em Attention Head}          & n/a & 8 \\\hline
\end{tabular}
}
\vspace{-6mm}
\end{table}

\vspace{-5mm}
\section{Experiments}
\vspace{-3mm}
We select 12 languages based on their richness of resources and their writing script, as shown in Table~\ref{tab:langs}. For example, \cblue{we see Russian and Kazakh share the same Cyrillic script where the latter is a low-resource language.} \cblue{For the model input/output text tokenization, we use SentencePiece tokenizer model (SPM)\cite{kudo2018sentencepiece} with a vocabulary size of 20,000. Our ablation study with varying SPM vocabulary sizes (omitted due to space constraints) shows little performance improvement beyond that. To note, unless specified in experiment results, LSTM based Seq2Seq architecture is our default ITN model as described in Section \ref{sec:model_arch}.
As choice of NMT models for our experiments, we use (a) Opus-MT \cite{TiedemannThottingal:EAMT2020} with EasyNMT library support, (b) MetaNMT, an in-house version of NMT with comparable performance to NLLB \cite{nllb2022}. Also, unless specified, we use MetaNMT for data augmentation as it has better BLEU scores (see respective papers) and ITN performance impact (see Table \ref{tab:ablation}).}
\vspace{-2mm}
\begin{table}[h]
\centering
\vspace{-4mm}
\caption{12 languages selected for experiments.}
\vspace{1mm}
{ \scriptsize
  \begin{tabular}{c|c|c}
    \hline
    \textbf{Resources} & \textbf{Latin Script} & \textbf{Non-Latin Script} \\
    \hline
   	\multirow{2}{*}{High} & Italian [\textit{it}], French [\textit{fr}], Spanish [\textit{es}], & Russian [\textit{ru}], Greek [\textit{el}]\\
   	& English [\textit{en}], Turkish [\textit{tr}], German [\textit{de}] & \\
   	\hline
   	Low & Icelandic [\textit{is}], Afrikaans [\textit{af}] & Tamil [\textit{ta}], Kazakh [\textit{kk}]\\
   	\hline
  \end{tabular}
}
\vspace{-8mm}
  \label{tab:langs}
\end{table}

\vspace{-2mm}
\subsection{Dataset}
\vspace{-2mm}
We use the OpenSubtitles\cite{lison2016opensubtitles2016} and TED2020\cite{reimers-2020-multilingual-sentence-bert} datasets from OPUS\footnote{https://opus.nlpl.eu/} as our training data. To be specific, only the written English texts from OPUS are used in our method. The spoken-written pairs for training are generated using the pipeline in Fig.~\ref{fig:pipeline}.

\cblue{We evaluate our models on two datasets: (a) Dictation testset: Human annotated 6,810 spoken-written conversational text pairs in English containing diverse ITN entities in mixed proportions. We apply the Case B strategy described in Fig \ref{fig:itn_eval} and  Section~\ref{sec:itn_eval} to generate evaluation data for non-English languages. (b) Caption testset: Mostly containing mathematical expressions, measures, metrics, phone numbers collected from \textit{audios with uploaded caption}. This dataset has  [\textit{en}]:22332, [\textit{es}]:21216, [\textit{fr}]:27300, [\textit{it}]:14939, [\textit{de}]:5960 spoken-written text pairs for respective languages.}

\vspace{-4mm}
\subsection{ITN Evaluation}
\label{sec:itn_eval}
\vspace{-1mm}
For the evaluation of our ITN model, we \cblue{mainly} focus on the numerical entities in the text. The evaluation is straightforward when ground-truth target language spoken-written pairs are available. But the lacking of proper human annotated spoken-written pairs for mid/low resource languages motivates us to propose an approach to measure the model performance on the numerical entities shown in Fig.~\ref{fig:itn_eval}.

\vspace{-4mm}
\subsubsection{ITN normalized accuracy}
\vspace{-1mm}

\begin{figure}[t]
    \vspace{-8mm}
    \includegraphics[width=\linewidth]{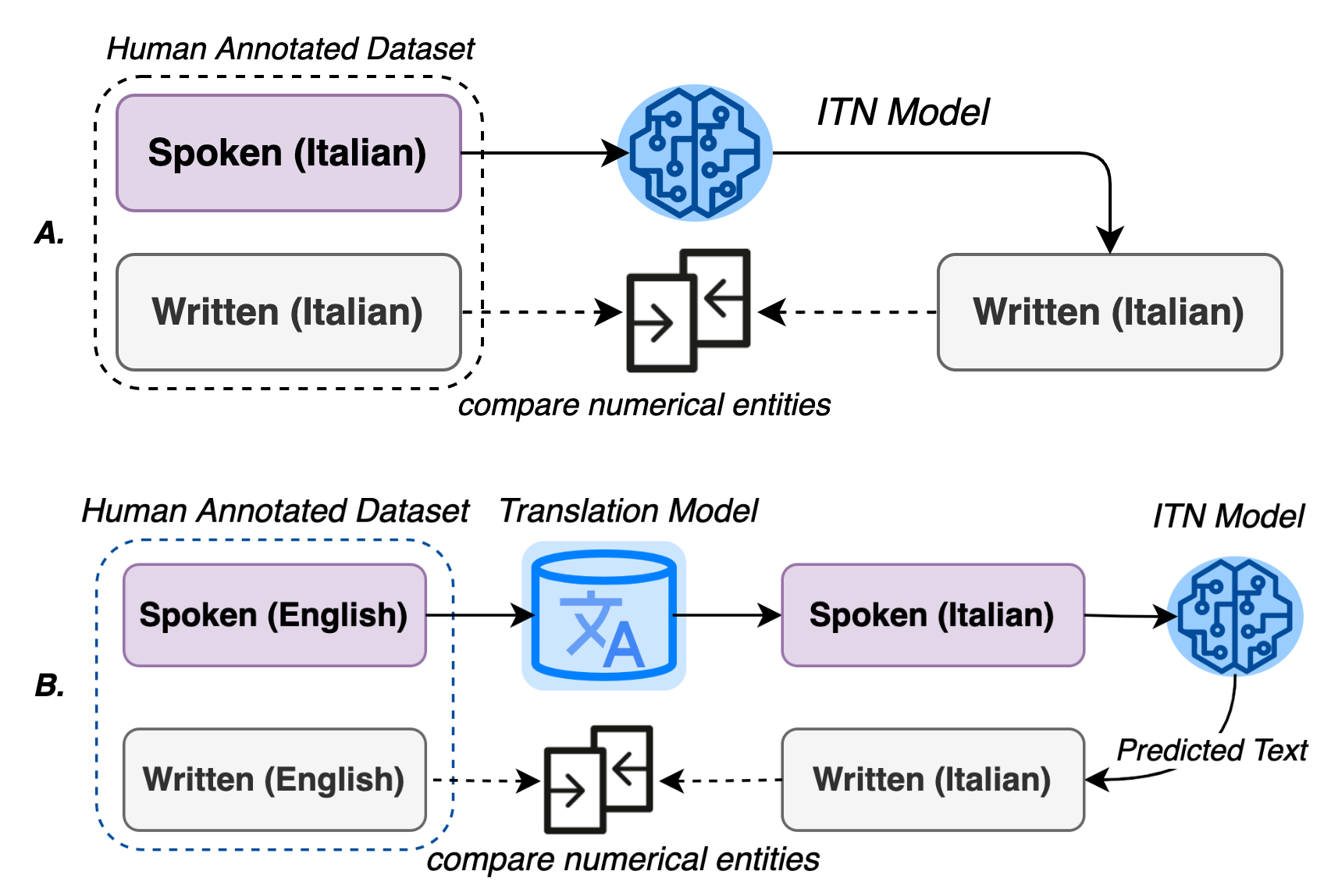}
    \vspace{-7mm}
    \caption{ITN evaluation strategy with case A: Target language evaluation dataset is available; Case B: Target language evaluation dataset is  not available. We use translation model to prepare spoken target texts. In both cases, we compare numerical entities output from the ITN model.
    }
    \label{fig:itn_eval}
    \vspace{-5mm}
\end{figure}

In Fig.~\ref{fig:itn_eval}, Case A shows the evaluation approach when target language spoken-written pair dataset is available. \cblue{We apply Case B where} target language pair dataset is not available, we translate the spoken form text from English human annotated dataset to target language. We then apply the trained ITN model to obtain written form text in target language. Thereafter, we verify the digit in the written form is the same as the original one in English. If they are the same, we count it as a correct instance, if not, we count it as a incorrect one. To note, \cblue{translation models may output written form in target language for spoken source input, we discard them and only evaluate the ITN model on correctly translated spoken form texts.}
\crblue{We use normalized accuracy to measure the performance of our ITN model expressed as the fraction of correct prediction over all the ITN entities.}


Table~\ref{tab:eval_errors} shows ITN entities in \textbf{bold}, the correct spoken and written form translated entities are in {\color{blue}\textbf{blue}} and errors are colored with {\color{red}\textbf{red}} with correct form in \textbf{(parenthesis)}. When comparing written form English and written form in target language, we take care of special cases; for example see the following examples:
\vspace{-2mm}
\begin{itemize}[leftmargin=*]
    \item \textbf{12/24 hour conversion}: Use of 24 hours system in French and other target language; \textit{ 1:30 p.m. [\textit{en}]} vs. \textit{13h30 [\textit{fr}]}.\\ \vspace{-7mm}
    \item \textbf{Accommodating Zeros}: Use alternate magnitudes; e.g., \textit{24,000 [\textit{en}]} vs. \textit{24 mille [fr]}.\\ \vspace{-7mm}
    \item \textbf{Small cardinal}: Disambiguate use of spoken/written form for small cardinals (1-9) based on target language preferences; e.g., \textit{two children [\textit{en}]} vs. \textit{2 enfants [fr]}.\\ \vspace{-7mm}
    \item \textbf{Separators}: Some languages like French use comma(,) for decimal point. Also, there are languages that use 2 or 3 digit separators with either  comma, space or period; e.g., \textit{25,000.00 [\textit{en}]} vs. \textit{25 000,00 [fr]} vs. \textit{25.000,00 [de]}.\\ \vspace{-7mm}
\end{itemize}

\begin{table}[t]
\centering
\vspace{-8mm}
\caption{Examples of errors for ITN evaluation. The second row is an example of ITN error while the third row is an example of NMT error.}
\vspace{1mm}
\label{tab:eval_errors}
{\scriptsize
\begin{center}
\vspace{-3mm}
\begin{tabular}{l|l}
\hline
\textbf{Spoken}  & \textbf{Written} \\ \hline
\begin{tabular}[c]{@{}l@{}}[\textit{en}] I found out that \textbf{nine} out of \textbf{ten} \\ statistics are wrong. \end{tabular}             & \begin{tabular}[c]{@{}l@{}}[\textit{en}] I found out that \textbf{9} out of \textbf{10} \\ statistics are wrong. \end{tabular}  \\
\begin{tabular}[c]{@{}l@{}}[\textit{fr}] J'ai découvert que {\color{blue}\textbf{neuf}} statistiques \\sur {\color{blue}\textbf{dix}} sont fausses.
\end{tabular}             & \begin{tabular}[c]{@{}l@{}}[\textit{fr}] J'ai récemment découvert que\\ {\color{red}\textbf{neuf}} \textbf{(9)} statistiques sur {\color{blue}\textbf{10}} sont fausses \end{tabular}  \\ \hline
\begin{tabular}[c]{@{}l@{}}[\textit{en}] Dad's surprise \textbf{sixtieth} is on this\\ Saturday. Arrive before \textbf{six PM}. \end{tabular}             & \begin{tabular}[c]{@{}l@{}}[\textit{en}] Dad's surprise \textbf{60th} is on this \\Saturday. Arrive before \textbf{6 PM}. \end{tabular}  \\
\begin{tabular}[c]{@{}l@{}}[\textit{fr}] La {\color{blue}\textbf{soixantième}} surprise de papa \\a lieu ce samedi. Arrivée avant {\color{red}\textbf{18h}}\\ \textbf{(dix-huit heures)}. \end{tabular}             & \begin{tabular}[c]{@{}l@{}}[\textit{fr}] La {\color{blue}\textbf{60ème}} surprise de papa \\a lieu ce samedi. Arrivée avant {\color{blue}\textbf{18h}}. \end{tabular}  \\ \hline
\end{tabular}
\end{center}
\vspace{-4mm}
}
\vspace{-6mm}
\end{table}

\vspace{-6mm}
\subsubsection{ITN entity translation accuracy}
\vspace{-1mm}
In order to evaluate the NMT model translation accuracy on ITN entities, we propose the ITN entity translation accuracy.
Here we compare whether the numerical entities from translations produces consistent spoken/written form output w.r.t. input text form.
\cblue{An example is provided in the last row of Table~\ref{tab:eval_errors}.}
Furthermore, we apply back-translation strategy with the same/other NMT and compare it with original text ensuring quality dataset from translation.


\vspace{-5mm}
\subsection{Results and Analysis}
\vspace{-2mm}
\cblue{We show the performance of monolingual baseline models and multilingual models on the \cblue{Dictation testset} in Table~\ref{tab:multilingual}.} Since the multilingual models have the same size with the monolingual models, these results are showing the impact of training data. For the high-resource languages, the performance gain of multilingual model is quite limited, and even worse performance \cblue{is obtained} (e.g., [\textit{es}]). 
On the other hand, we observe that the multilingual models significantly improve the performance on low-resource languages (except Tamil [\textit{ta}] since it is using a different script). For example, the normalized accuracy of monolingual model on Kazakh [\textit{kk}] is only 0.03\%, and the multilingual models could achieve the performance of 37.69\%. Thus, we could say that adding training data from the same script can improve the model performance on low-resource languages. As a reference, we also provide the ITN entities translation accuracy in the last column of Table~\ref{tab:multilingual}. The higher the accuracy, the more ITN entities are evaluated in the \cblue{Dictation testset}.

Similarly, Table~\ref{tab:base1} shows the \cblue{performance of} monolingual baseline models \cblue{and the 12-language model} on the \cblue{Caption Testset}. This is computed by \cblue{the case A strategy in Fig.~\ref{fig:itn_eval}.} \cblue{The 12-language ITN model is able to perform better accuracy on [\textit{en, es, de}] while maintain competitive accuracy on [\textit{fr, it}].}

From these two tables, we can see that the normalized accuracy is much higher on the \cblue{Dictation testset} than on the \cblue{Caption Testset}. \cblue{We find that there is a domain mismatch in training and testset, as we find a lot of complex math expressions in the Caption testset (e.g., $2+4+2+7+4=19$)  while none is found in the open subtitles training set.}

\begin{table}[t]
\vspace{-8mm}
  \caption{Normalized accuracy of monolingual and multilingual models on the \cblue{Dictation testset}.}
  \label{tab:multilingual}
  \vspace{1mm}
  \scalebox{0.77}{
  \begin{threeparttable}
  \begin{tabular}{c|c|ccc||c}
    \hline 
    \textbf{Lang.$^\ddagger$} & \textbf{Monolingual} & \textbf{3-lang.$^\ddagger$} & \textbf{6-lang.$^\ddagger$} & \textbf{12-lang.$^\ddagger$} & \textbf{ITN trans.$^\ddagger$ acc.$^\ddagger$}\\
    \hline
   	\textit{es} & \textbf{79.15\%} & 78.09\% & 76.80\% & 75.17\% & 91.34\%\\
   	\textit{fr} & 62.35\% & \textbf{62.99\%} & 60.98\% & 60.07\% & 62.81\%\\
   	\textit{it} & 70.71\% & \textbf{71.42\%} & 69.96\% & 69.87\% & 76.02\%\\
   	\textit{en} & 71.73\% & - & \textbf{72.75\%} & 71.96\% & - \\
   	\hline
   	\textit{ru} & \textbf{68.39\%} & - & 64.66\% & 66.33\% & 82.86\%\\
   	\textit{kk}\tnote{\textdagger} & 0.03\% & - & \textbf{37.69\%} & 32.41\% & 99.63\%\\
   	\textit{tr} & \textbf{60.07\%} & - & - & 53.95\% & 46.19\%\\
   	\textit{de} & \textbf{68.24\%} & - & - & 63.74\% & 61.67\%\\
   	\hline
   	\textit{el} & \textbf{66.84\%} & - & - & 65.29\% & 64.64\%\\
   	\textit{is}\tnote{\textdagger} & 48.50\% & - & - & \textbf{61.75\%} & 99.36\%\\
   	\textit{af}\tnote{\textdagger} & 29.21\% & - & - & \textbf{50.51\%} & 96.45\%\\
   	\textit{ta}\tnote{\textdagger} & 25.63\% & - & - & \textbf{27.30\%} & 99.74\%\\
   	\hline
  \end{tabular}
  \begin{tablenotes}
    \item[$^\ddagger$]lang., trans., and acc. stands for languages, translation, and accuracy.
    \item[\textdagger] low resource languages.
  \end{tablenotes}
  \end{threeparttable}}
  \vspace{-5mm}
\end{table}

\begin{table}[t]
{
\vspace{-2mm}
\small
  \caption{\cblue{Normalized accuracy of} monolingual and {12-language} \cblue{model} on the \cblue{Caption testset}.}
  \vspace{.5mm}
  \label{tab:base1}
  \scalebox{0.87}{
  \begin{threeparttable}
  \begin{tabular}{c|c c c c c}
    \hline 
    Language & \textit{en} & \textit{es} & \textit{fr} & \textit{it} & \textit{de}\\
    \hline
    Monolingual & 63.70\% & 64.51\% & \textbf{55.24\%} & \textbf{57.57\%} & 48.10\% \\
    12-language & \textbf{64.74\%} & \textbf{65.58\%} & 54.90\% & 56.77\% & \textbf{50.19\%} \\
   	\hline
  \end{tabular}
  \end{threeparttable}}
  \vspace{-5.5mm}
}
\end{table}

\begin{table}[!t]
{\small
\vspace{-1mm}
  \caption{Normalized accuracy when comparing of architecture and translation tools on 3-langs ([\textit{es}], [\textit{fr}], [\textit{it}]) model and SPM token size of 20,000 on the \cblue{Dictation testset}.}
  \label{tab:ablation}
  \begin{center}
  \vspace{-2mm}
  \begin{tabular}{c|c|ccc}
    \hline
    \textbf{Arch.} & \textbf{NMT} & \textit{es} & \textit{fr} & \textit{it} \\
    \hline
    Seq2Seq & MetaNMT & \textbf{78.09\%} & \textbf{62.99\%} & \textbf{71.42\%} \\
   	Seq2Seq & Opus-MT & 71.11\% & 60.03\% & 55.89\% \\
   	Transformer & MetaNMT & 72.55\% & 57.27\% & 64.76\% \\
   	\hline
  \end{tabular}
  \end{center}
}
  \vspace{-6mm}
\end{table}

\begin{table}[!t]
{\small
\vspace{-4mm}
  \caption{Comparison of standard rule-based ITN system and data-driven ITN with/without data augmentation for English.}
  \label{table:en_compare}
  \vspace{-3mm}
\begin{center}
\scalebox{0.9}{
  \begin{tabular}{c|c|c|c}
    \hline
    & & \textbf{DD-ITN without}  & \textbf{DD-ITN with}  \\
    & \textbf{Rule-based ITN}  & \textbf{ data augmentation} & \textbf{data augmentation} \\ \hline
    \textit{en}& 63.20\%  & 51.40\% & \textbf{86.20\%}    \\ \hline
  \end{tabular}
}
\end{center}
\vspace{-9mm}
}
\end{table}


Finally, Table~\ref{tab:ablation} shows the ITN normalized accuracy when using a different model architecture and different machine translation model \cite{TiedemannThottingal:EAMT2020}; shows that the LSTM-based Seq2Seq model is better than the Transformer model, and MetaNMT is better than the Opus-MT for our task. \crblue{Table \ref{table:en_compare} compares standard rule-based ITN with DD-ITN and illustrates the impact of utilizing data-driven ITN modeling with our enhanced text normalization data augmentation for English on the Google TN evaluation dataset \cite{google_tn} with more than $70K$ ITN entities.}\\
\vspace{-8mm}
\section{Conclusion}
\vspace{-3mm}
This work investigates the effectiveness of language agnostic data-driven ITN model. With the same model size, a 12-languages ITN model can significantly improves the normalized accuracy of low resource languages while maintain reasonable performance for high resource languages. Also, we explore the architecture and the machine translation model used in our framework and found that Seq2Seq model with MetaNMT is our best system. \cblue{Our future work will include designing better architecture for on-device deployment, evaluation with clean and native language dataset, adapt data augmentation to support native language nuances/variation, and expanding to more languages.}



\bibliographystyle{IEEEbib}
\bibliography{main}

\crblue{
    \newpage
    \section{Appendix}
\subsection{Data Augmentation with enhanced TN}
Unlike a conventional written to spoken text normalization system, our enhanced text normalization data augmentation system for data-driven ITN is capable of generating much diversified spoken forms by introducing almost all possible spoken variations to the written forms as shown in Table \ref{table2}. Our rule-based enhance TN system supports cardinal, ordinal, decimal, fraction, measures, money, date, time and telephone entities. This system performs a series of steps for each written text as input; pictorially depicted in Figure \ref{fig:pipeline} and describe as follows. 
\begin{itemize}
    \item[a.] \textit{Pick:} From the text corpus pick up sentences with numerical ITN entities and discard other sentences. \\ \vspace{-6mm}
    \item[b.] \textit{Categorize and Segment:} Extract text chunks that matches our carefully prepared regular expressions for each type of ITN entities in a predefined order. First, time and date entities segments are collected, followed by measures, currency, abbreviations. Then we find fractions, decimals, ordinals, phone numbers and cardinal in respective order. \\ \vspace{-6mm}
    \item[c.] \textit{Verbalize:} Entities matching text segment are then cleaned, formatted and normalized with digit and spoken words to its closest verbalized form. For examples, {\em Time: 12:45 $\rightarrow$ 12 hours 45 minutes.}; {\em Date: 12/31/2022 or (31-12-2022) $\rightarrow$ 31 December 2022}; {\em Measures: 10K lb $\rightarrow$ 10000 lb.; 207.6 kmps $\rightarrow$ 207.6 kilometers per second.;} {\em 2 kg $\rightarrow$ two kilogram, two kilos, two kilo}, etc.  \\ \vspace{-6mm}
    \item[d.] \textit{Numeric Normalization:} With verbalized text, we apply  our rewrite mapping rules with core logic by selecting all possible combination of digits in order (and apply recursively, if required) to generate $N$ spoken form texts for corresponding written text.\\ \vspace{-6mm}
    \item[e.] \textit{Rewrite:} Finally, a rewrite module replaces the written-form text in the original sentence with $N$ generated diverse spoken forms. \\ \vspace{-6mm}
\end{itemize}
\vspace{-2mm}
\begin{table}[h]
\vspace{-3mm}
\caption{Examples of generated spoken form using conventional TN system and our enhanced TN system}
\label{table2}
{\scriptsize
\begin{center}
\vspace{-2mm}
\begin{tabular}{l|l|l}
\hline
\textbf{Input} & \textbf{Conventional TN}  & \textbf{Our Augmentation System} \\ \hline
123            & \begin{tabular}[c]{@{}l@{}}one hundred \\ twenty three\end{tabular}    & \begin{tabular}[c]{@{}l@{}}one hundred twenty three\\ one twenty three\\ one hundred and twenty three\\ one two three\end{tabular}                                                                                                                                                          \\ \hline
\$123          & \begin{tabular}[c]{@{}l@{}}one hundred  \\ twenty three dollars\end{tabular} & \begin{tabular}[c]{@{}l@{}}one hundred twenty three dollars\\ one hundred twenty three dollar\\ one twenty three dollars\\ one twenty three dollar\\ one hundred and twenty three dollars\\ one twenty three dollars zero cents\end{tabular} \\ \hline
123g           & \begin{tabular}[c]{@{}l@{}}one hundred \\  twenty three grams\end{tabular}    & \begin{tabular}[c]{@{}l@{}}one hundred twenty three grams\\ one hundred twenty three gram\\ one twenty three grams\\ one twenty three gram\\ one hundred and twenty three grams\\ one hundred and twenty three gram\\ one two three grams\end{tabular}                 \\\hline
\end{tabular}
\end{center}
}
\vspace{-12mm}
\end{table}

\subsection{Non-ITN Accuracy:}\vspace{-2mm}
To minimize errors related to entities that are not part of the ITN (i.e., non-ITN entities) from our sequence-to-sequence based model, we have implemented the following measures:
\begin{itemize} \vspace{-3mm}
    \item[a.] \textit{Cleansing at source:} We filter out spoken-written pairs with non-zero or very low word-error-rate for non-numerical segments from our training data. It is important to note that dividing large texts into small segments helps in achieving highly accurate alignment and thus aids in the removal of errors at the source.\\ \vspace{-6mm}
    \item[b.] \textit{Selective Inference:} Recent ASR systems can provide hints about the entity type of a text segment, which is a simpler task than performing the ITN post-processing with ASR. Additionally, it is also easy to train a standalone non-ITN text classifier using a sequence encoder model with our augmentation dataset. We apply selective inference of numerical entities to ensure that non-ITN entities are not translated.\\ \vspace{-6mm}
    \item[c.] \textit{Accuracy Metric:} We have also performed evaluations of non-ITN accuracy for languages such as Spanish, French, Italian, and German, and found that the accuracy is greater than 98\% on a human-annotated dataset.\\ \vspace{-8mm}
\end{itemize}

\vspace{-2mm}
\subsection{Comparison of ITN systems} \vspace{-2mm}
Here we present experiment and analysis of rule-based ITN system and data-driven ITN with/without our enhanced text normalization data augmentation system for English language on evaluation set of Google's Text Normalization dataset \cite{google_tn}.

The rule-based ITN model, which relies on manually crafted grammars, is difficult to maintain and prone to errors due to its complexity. Additionally, it performs poorly on entities that require contextual information. In contrast, our end-to-end data-driven approach eliminates the need for complex rules, making it easier to scale to new languages and perform better on entities requiring contextual information. From the result, it shows that data driven ITN together with augmentation perform significantly better than rule based ITN, which achieved 36.4\% improvement on overall entity accuracy. We see significant improvement on Decimal, Measure and Time entities.\\
\begin{table}[h]
\vspace{-8mm}
{\scriptsize
\caption{Comparison of normalized accuracy for rule-based ITN system and data-driven ITN with/without enhanced TN data augmentation for English language.}
\label{table:compare}
\begin{center}
\vspace{-2mm}
\begin{tabular}{l|r|c|c|c} \hline
     &   & \textbf{Rule-based} & \textbf{Data-driven ITN}  & \textbf{Data-driven ITN}  \\
  \textbf{Class}     & \textbf{Size} &\textbf{ ITN} & \textbf{without data} & \textbf{with data} \\
  &  & & \textbf{augmentation} & \textbf{ augmentation} \\\hline
Cardinal  & 10000 & 92.00\%          & 55.00\%                               & \textbf{97.00\%}                            \\
Ordinal   & 10000 & 82.87\%          & 50.90\%                               & \textbf{97.50\% }                           \\
Decimal   & 10000 & 2.20\%           & 53.10\%                               & \textbf{90.80\%   }                         \\
Measure   & 10000 & 46.70\%          & 42.79\%                               & \textbf{88.20\%}                            \\
Telephone & 4024  & 93.20\%          & 88.00\%                               & \textbf{93.80\% }                           \\
Digit     & 5442  & 77.60\%          & 56.80\%                               & \textbf{90.60\% }                           \\
Time      & 1159  & 36.10\%          & 35.40\%                               & \textbf{73.60\% }                           \\
Date      & 10000 & 84.50\%          & 46.00\%                               & \textbf{89.40\%  }                          \\
Money     & 10000 & 53.80\%          & 34.90\%                               & \textbf{55.40\%}                            \\ \hline
\textbf{Overall}   & \textbf{70625} & \textbf{63.20\%}          & \textbf{51.40\%   }                            & \textbf{86.20\%}    \\ \hline                    
\end{tabular}
\end{center}
}
\end{table}

}

\end{document}